\documentclass[11pt,a4paper]{article}

\usepackage{url}
\usepackage[T1]{fontenc}
\usepackage[dvipsnames]{xcolor}
\usepackage{multirow}
\usepackage{tablefootnote}
\usepackage{times}
\usepackage{latexsym}
\usepackage{amsmath}
\usepackage{amssymb}
\usepackage{graphics}
\usepackage{graphicx}
\usepackage{hyperref}
\usepackage{dsfont}
\usepackage{xspace}
\usepackage{verbatim}
\usepackage{xcolor}
\usepackage{soul}
\usepackage{tabularx}
\usepackage{booktabs}
\usepackage{caption}
\usepackage{geometry}
\usepackage{comment}
\usepackage{enumitem}
\usepackage{subfigure}

\usepackage{todonotes}

\newcommand{\name}{\textsc{StyleMC}}

\def\equationautorefname~#1\null{%
  Equation~(#1)\null
}
\def\itemautorefname~#1\null{(#1)\null}
\def\sectionautorefname~#1\null{\S#1\null}
\def\subsectionautorefname~#1\null{\S#1\null}
\def\subsubsectionautorefname~#1\null{\S#1\null}
\usepackage[acceptedWithA]{tacl2021v1}

\usepackage{xspace,mfirstuc,tabulary}

\newif\iftaclinstructions
\taclinstructionsfalse 
\iftaclinstructions
\renewcommand{\confidential}{}
\renewcommand{\anonsubtext}{(No author info supplied here, for consistency with
TACL-submission anonymization requirements)}
\newcommand{\instr}
\fi

\iftaclpubformat 

\else

\fi

\title{Learning to Generate Text in Arbitrary Writing Styles} 

\author{Aleem Khan, Andrew Wang, Sophia Hager \and Nicholas Andrews \\
Department of Computer Science,
        Johns Hopkins University\\
        \texttt{\{akhan141,awang116,shager2,noa\}@jhu.edu}}

\date{}

\begin{document}

\maketitle
\begin{abstract}
Prior work in style-controlled text generation has focused on tasks such as emulating the style of prolific literary authors, producing formal or informal text, and mitigating toxicity of generated text. Plentiful demonstrations of these styles are available, and as a result modern language models are often able to emulate them, either via prompting or discriminative control. However, in applications such as writing assistants, it is desirable for language models to produce text in an \emph{author-specific} style on the basis of a potentially small writing sample.
For example, someone writing in a particular dialect may prefer writing suggestions that retain the same dialect.
We find that instruction-tuned language models can struggle to reproduce author-specific style demonstrated in a prompt. 
Instead, we propose to guide a language model to generate text in a target style using contrastively-trained representations that capture stylometric features.
Our approach (\name) combines an author-adapted language model with sequence-level inference to improve stylistic consistency, and is found to be effective in a variety of conditions, including unconditional generation and style transfer. Additionally, we find that the proposed approach can serve as an effective anonymization method, by editing a document to mask authorship while preserving the original meaning.
  
\end{abstract}

\section{Introduction}\label{sec:intro}
We consider the problem of generating text in the style of an arbitrary author on the basis of a small writing sample, on the order of a few hundred words. Although instruction-tuned language models (LM) have demonstrated the ability to emulate a variety of writing styles via prompting \cite{deshpande2023toxicity}, particularly when a given style is well-represented in the training data \cite{krishna2020reformulating},
we find that performance is less consistent in our few-shot setting, with recent large LMs such as GPT-3.5 performing worse than its previous generations.
A separate challenge is that large LMs can be computationally prohibitive in certain applications, such as on-device deployment where privacy-preserving personalized generation may be needed.

\begin{figure}[t]
    \centering
    \includegraphics[scale=0.67,trim=1.9cm 0.65cm 12.0cm 2.9cm,clip=true]{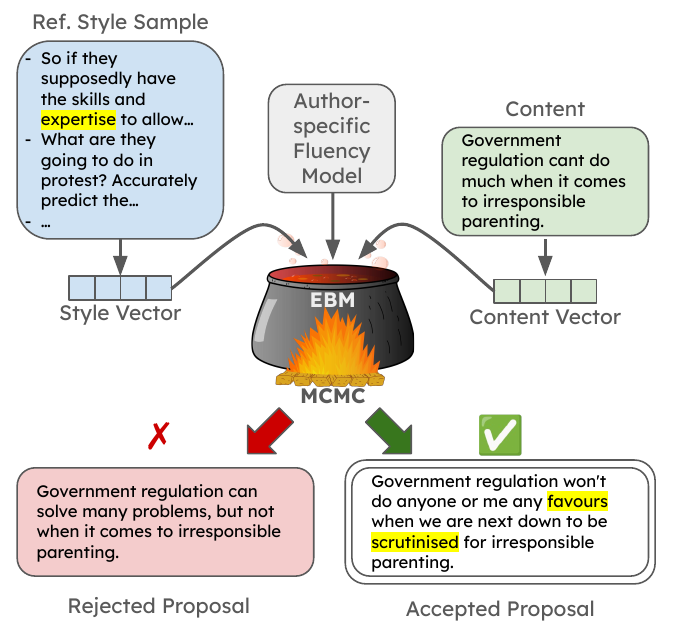}
    \caption{An overview of \name~for style transfer. Using MCMC, we generate text based on a few samples of the target style, an author-specific fluency model (\S\ref{sec:ebm}), and the original content. \textbf{Our approach reproduces salient features of an author's style while preserving meaning.} In the (real) example above, accepted samples exhibit characteristics of British English (yellow), matching the characteristics of the target style.}
    \label{fig:intro-figure}
\end{figure}

Prior work in controllable text generation has primarily focused on  categorical target attributes such as sentiment, formality, and topic, for which a number of techniques have been proposed \cite{prabhumoye-etal-2018-style, sudhakar2019transforming, yang2021fudge}---we discuss related work in more detail in~\autoref{sec:related_work}. 

However, author-specific textual styles cannot be summarized using a closed set of binary or categorical attributes, since authors may be characterized by unique combinations of stylometric features. Such features may include dialect, use of emojis, punctuation and capitalization usage, as well as less obvious features such as syntactic preferences and use of white space.
Since it is difficult even for forensic linguists to characterize an author's style, we propose to guide generation using
contrastively-trained representations that extract stylistic attributes from a given writing sample as a dense vector feature.\footnote{A conceptually similar approach is used in certain voice synthesis systems, in which \emph{speaker} representations guide qualities of the generated speech~\cite{fang2019speaker,ao2021speecht5}.} 

Discriminative control methods generate text with prescribed attributes guided by a classifier evaluating the degree to which the text satisfies the target attribute, typically with a tunable hyper-parameter balancing the \emph{fluency} of the generated text with control success~\cite{dathathri2019plug}.
However, human writing is characterized by ``dips'' into low-probability regions, unlike samples from LMs which produce likely tokens at each step~\cite{gehrmann2019gltr}.
Thus, the objective of achieving fluent generations according to a generic LM will in general be in tension with the goal of matching an author's style, which may be characterized by such unlikely token choices.
To overcome this challenge, we propose \name, a novel approach which combines a style-controlled autoregressive language model and a discriminative objective which aims to ensure stylistic consistency at the sequence level.

To guide a pre-trained LM towards a target style, we generalize future discriminators~\cite{yang2021fudge} to \emph{regression} for a target style representation. Simply put, our approach involves re-scoring the predictive distribution of an existing LM using a lightweight model that assigns higher likelihood to tokens that are predicted to better adhere to the target style vector. The resulting \emph{author-specific} LM---the composition of a pre-trained model and a lightweight regressor---is then used as a fluency scorer for a discriminative model which captures stylistic consistency at the document-level. Our discriminative control framework adopts a  product-of-experts energy parametrization~\cite{hinton-2004-training,mireshghallah-etal-2022-mix}, which enables \textit{style transfer} through the inclusion of meaning preservation terms (\autoref{sec:style-transfer-section}). 

In summary, \name~enables both style-controlled generation and style transfer using a pre-trained LM, without further fine-tuning. Our recipe calls for two main ingredients: a style representation (\autoref{sec:style_rep}) and unlabeled data to fit a lightweight re-scoring model. Since style representations are effective in various domains and unlabeled data is generally easy to come by, our approach is quite widely applicable. We conduct an extensive experimental evaluation of the proposed approach in~\autoref{sec:style-control-experiments}, \autoref{sec:style-transfer-section}, and~\autoref{sec:added-experiments}, finding that:

\begin{itemize}[nosep,leftmargin=*]
\item \name~is proficient in style generation tasks (including style transfer), outperforming large language models such as GPT-4 that were prompted to use in-context demonstrations of the target style.

\item Interpolating between two style vectors and generating text at intermediate points using our approach yields interpretable results, considering the rate of capitalization and punctuation usage. This result suggests that (a) our control vectors capture intuitive stylistic features; and (b) that the proposed approach can successfully reproduce those features in generated text at the expected rate. 

\item Our proposed style transfer approach can be adapted to serve as an effective author \emph{anonymization} technique, defeating authorship attribution while preserving meaning.
\item In a zero-shot setting, samples from the proposed approach are harder to detect as being machine-generated than other LLMs, which we attribute to a greater ability to mimic human writing style.
\end{itemize}
\paragraph{Reproducibility} We release reference model implementations, checkpoints, datasets, and experiment scripts. Commodity hardware (e.g., a single GPU) is sufficient to reproduce most of our results.

\section{Preliminaries}\label{sec:prelim}

\subsection{Problem statement}
 We consider both sequence generation (\autoref{sec:fudge}) and sequence-to-sequence generation (\autoref{sec:ebm}), where in both cases our objective is to produce text ($x$) in a target style while satisfying other criteria, such as diverse outputs in the case of language modeling and meaning preservation in the case of style transfer. We assume a few-shot setting where the target style is specified by a writing sample $y = (y_1, y_2, \ldots)$ exhibiting the desired stylistic attributes. In our experiments, we focus on the case where each $y_1, y_2, \ldots$ correspond to a short documents (e.g., social media comments), and we are interested in reproducing the underlying author's specific writing style. We emphasize the difficulty of this task, stemming not only from the few-shot setting, but also the fact that stylometric features comprise a sparser signal than other more evident textual attributes like sentiment.\footnote{Statistical authorship analysis often assumes access to large corpora by the candidate authors, as in the seminal work by~\citet{mosteller1963inference}. In contrast, we extract features from a relatively small number of short documents, on average comprising 68 words each.}

For sequence generation, we produce text by sampling from a pre-trained LM $p$ conditioned on $y$. In the case of instruction-tuned LMs, $y$ will be paired with an appropriate prompt to elicit the desired output; we discuss prompting strategies in more detail in~\autoref{sec:prompted_lms}.
In sequence-to-sequence generation, we are additionally given initial text $x^{(0)}$ that we wish to revise to be closer to the target style $y$, while keeping other properties of $x^{(0)}$ constant, such as semantic meaning. 
Rather than condition on $y$ directly as done in the prompting approach, we propose instead using a discriminative feature extractor $f$ to capture stylistic properties of $y$, which are used to guide generation. This is a distinguishing characteristic of our approach, since much prior work in controllable text generation has focused on classifiers (e.g., sentiment polarity) and prompting strategies to guide generation. We discuss the feature extractor in more detail next.

\subsection{Author style representations}\label{sec:style_rep}

As previously mentioned, author-specific style is difficult to characterize even for forensic linguists, which poses challenges both for control and for evaluation. However, recent work has leveraged the availability of large corpora of writings by anonymous authors to learn stylistic representations. Such representations have been found to be effective at discriminating between authors by characterizing writing style~\cite{luar-style}.  In this work, we consider two different representations, both trained for surrogate tasks of authorship prediction. 
\paragraph{Control} To guide generation, we adapt the model proposed by~\citet{luar}. Specifically, we estimate a representation $f$ on the basis of a large collection of anonymous writing samples. Our training dataset consists of one million Reddit users, each contributing at least 100 comments~\cite{baumgartner2020pushshift,khan-etal-2021-deep}. The unique account labels enable supervised contrastive training, encouraging features $f(x)$ and $f(x')$ to be similar when $x$ and $x'$ have the same author.\footnote{We use code provided by \citet{luar} at \texttt{https://github.com/LLNL/LUAR}.}
\paragraph{Evaluation} We use two models for evaluation which are both available publicly as pre-trained checkpoints (\autoref{sec:metrics}). The first is a further instance based on the recipe from~\citet{luar} trained on a larger corpus of 5 million authors, resulting in a more capable model than the one we use to guide generation. We also use a model proposed by~\citet{wegmann-etal-2022-author}, which is trained on different data using topic labels to attempt to produce representations that are less sensitive to topical similarity.
\section{Guiding generations towards a target style representation}\label{sec:control}

In this section, we describe \name, which aims to generate text $x$ satisfying various soft constraints, the most important of which is adherence to the style demonstrated in the few-shot example $y$. To reconcile the tension between fluency and author-specific style, we first show how to use a regression model to guide an LM to produce text for which $f(x)$ is close to $f(y)$ in expectation. Next, we show how the resulting author-specific LM can be incorporated in an energy-based  model (EBM) using a product-of-experts, which confers two advantages. First, the EBM is a non-autoregressive model which performs inference at the sequence-level; therefore, the distance between $f(x')$ and $f(y)$ can be directly evaluated to score candidate generations $x'$. Second, this framework makes it straightforward to introduce further experts to satisfy arbitrary additional preferences, such as meaning preservation in the case of style transfer.

\subsection{Few-shot language model adaptation}\label{sec:fudge}



An autoregressive LM conditioned on a control attribute $c$,
 \begin{equation*}\label{eqn:clm}
p(x \mid c) = \prod_{i=1}^n p(x_i \mid x_1, \ldots, x_{i-1}, c)
\end{equation*}
admits the following factorization of the likelihood according to Bayes' rule:
\begin{equation*}\label{eqn:factorization}
 p(x_i \mid x_{1:i-1}, c) \propto \underbrace{p(c \mid x_{1:i})}_{\text{Control}} \underbrace{p(x_i \mid x_{1:i-1}}_{\text{LM}}) .
\end{equation*} 
\citet{yang2021fudge} propose using maximum-likelihood estimation to fit $p(c \mid x_{1:i})$, namely the probability that the control attribute $c$ \emph{will hold in the future}, given the current prefix $x_{1:i}$. Such a model can be estimated on the basis of text paired with observed control attributes, and is then used during generation as a token-level re-scoring mechanism.

This approach affords a natural extension to continuous control by fitting a future \emph{regressor} $p(f(x) = \mathbf{c} \mid x_{1:i})$, where $f(x)$ is evaluated on the sequence and the model is conditioned on all prefixes of the same sequence, and therefore learns to predict the probability that a given prefix $x_{1:i}$ will adhere to the target style in the future.
To do so, we stipulate that control vectors $\textbf{c}$ are distributed according to a multivariate Normal density,
and parameterize $\mu$ and $\Sigma$ using neural networks with input $x_{1:i}$ and where $\Sigma$ constrained to be a diagonal convariance. 
Specifically, we employ a shared network $\textbf{z} = g_{\theta}(x_{1:i})$ for both $\mu$ and $\Sigma$:
\begin{align*}
 \mu &:= \text{MLP}_{\phi}(\textbf{z}) \\
 \Sigma &:= \text{diag}(\text{softplus}(\text{MLP}_{\eta}(\textbf{z})))
\end{align*}
The parameters $\Theta = (\theta, \phi, \eta)$ are optimized on the basis of a corpus consisting of text paired with corresponding control vectors $\{(\textbf{c}, x)\}_{i=1}^N$. In general, and in the experiments reported in this paper,  $g_\theta$ will have many fewer parameters than the LM being guided, in which case evaluating $p(\mathbf{c} \mid x_{1:i-1})$ during generation introduces  a relatively small additional computational burden. Implementation details for this architecture are outlined in \autoref{sec:style-details}.



The diagonal covariance matrix implies that each component of the control vector is independent. While previous work in contrastive learning has found that explicitly enforcing decorrelation to be necessary for such an assumption to be effective \cite{Tao2021ClusteringfriendlyRL}, we find that the control vectors we consider already satisfy this condition quite well.\footnote{In fact, we trained a model using the decorrelation objective and found the associated control vectors yielded no noticeable improvement in downstream decoding.} 

\vspace{5pt} \noindent \textbf{Optimization} For each training instance $(\textbf{c}, x)$, we create the augmented set consisting of all prefixes $(\textbf{c}, x_{1:1})$, $(\textbf{c}, x_{1:2})$, $\ldots$, $(\textbf{c}, x_{1:n})$. Note that the target $\textbf{c}$ is the same for each prefix, since the regressor is predicting whether the target control vector will be true for the full $x$ on the basis of the supplied prefix. The parameters $\Theta$ of the regression model are optimized to maximize the log-likelihood of the
observed control vectors. We found it effective to initialize $g_\theta$ using the same model that extracted the reference control vectors \footnote{We found that using this initialization resulted in a 1.4\% performance improvement over a random initialization.}, before fine-tuning $\Theta$ on the augmented data.

\subsection{A unified sequence-level model for style control and style transfer}\label{sec:ebm}

The proposed future regressor can be combined with any autoregressive LM to produce samples $x$ with stylistic features $f(x)$ close to the target $f(y)$ in expectation. However, autoregressive generation incrementally constructs the sample $x$, and therefore cannot directly use the feature-space distance between $f(x)$---based on the complete sample $x$---and the target $f(y)$, to guide generation. Additionally, to support tasks such as style transfer~(\autoref{sec:style-transfer-section}), it is necessary to impose additional constraints on generation such as meaning preservation.

To address these limitations, we employ our adapted LM as one of several experts in an EBM. Specifically, we parameterize the probability of a sequence $x$ given a target style $y$ as a product-of-experts~\cite{hinton-2004-training,du2020compositional},
\begin{align}\label{eqn:poe}
p(x \mid y) \propto e^{-\sum_i \alpha_i E_i(x, y)}
\end{align}
with experts $E_i$ corresponding to soft constraints; this model assigns higher probability to sequences $x$ which \emph{simultaneously} satisfy all constraints. Since evaluating the above probability requires an intractable sum over all possible sequences $x$, we resort to approximate inference (\autoref{sec:inference}).  We consider two settings in our experiments: style-controlled generation and style transfer. In both settings, we have found it straighforward to tune the weights $\mathbf{\alpha}$ using validation data, although we note that maximum-likelihood estimation could be used instead to avoid any manual tuning.

\paragraph{Style-controlled generation} Here we use only two experts. $E_1$ is an author-specific LM as described in the previous section, which evaluates the negative log-probability of $x$ under the author-adapted LM. In our implementation, we average the log-probabilities for each token rather than summing them to ensure sequence length does not skew energy scores. $E_2$ is an expert measuring sequence-level style similarity. Specifically, $E_2$ computes the distance between the style vector of $x$ and a target style control vector $f(y)$ via the negative angular similarity.To avoid noisy estimates for $f(x)$ when dealing with short text samples, we consider a candidates post within the context of other samples from the same author. Specifically, when revising a text sample $x_i$ from the writing sample $x = (x_1, x_2 ...)$, our EBM computes $E_2(x, y)$ as opposed to $E_2(x_i, y)$.

\vspace{5pt}\noindent \textbf{Controlled text revision} In the style transfer task, we additionally condition generation on an initial state $x^{(0)}$, and the objective is to modify $x_0$ to adhere to the style of $y$ while preserving the original meaning of $x^{(0)}$. To do so, we employ $E_1$ and $E_2$ as before, but introduce further experts that are functions of $x$ and $x^{(0)}$ and measure meaning preservation. We note that various options are possible for this purpose and our specific choices may not be optimal in all cases. In our experiments, we refer to $E_3$ as the measure of semantic similarity. 
To ensure that $x$ makes minimal revisions to $x^{(0)}$, we additionally add $E_4$ defined as the Hamming distance between $x$ and $x^{(0)}$, which was also employed by~\citet{mireshghallah-etal-2022-mix}.
\subsection{Inference}\label{sec:inference}

We frame the generation problem as finding an output $x$ which minimizes energy defined by \autoref{eqn:poe}. Although this problem is intractable, the Metropolis-Hastings (MH) algorithm can be used to obtain an approximate sample from the desired distribution. Differing from~\citet{goyal2021exposing}, we use T5 \cite{raffel2019exploring}, an encoder-decoder model trained with an in-filling objective, to obtain proposals for the sampling scheme.\footnote{We also experimented with a masked language model and found that the proposals were significantly lower quality than those produced by T5.} At each step in the procedure, a fixed number of tokens are masked which T5 in-fills with a variable number of tokens---possibly fewer than were masked---to generate a candidate state. 
In \autoref{sec:proposal_appendix} we vary the proposal model and masking scheme, finding that masking two tokens at a time, and infilling with T5-3B produced the best results.

In general, the state of the sampler may consist of more than one document, which together must adhere to the target style. At each step, we sample one of the documents $i$ for a MH update uniformly at random, and make proposals according to the proposal model conditioned on $x_i$, but evaluate $E_2(x, y)$ based on the entire sampler state. Thus, the energy function captures similarities on the entire state as opposed to a single document. We also run the sampler for a fixed number of steps (80 times the length of the sequence). For decoding, we record the intermediate states at each iteration, compute the energy for each state, and select the $argmax$ as our final output.

For style transfer (\autoref{sec:style-transfer-section}), we experiment with restrictions on our proposal distribution based on part-of-speech tags. We hypothesize that nouns most often do not characterize writing style and so can be fixed during our inference procedure without impacting the ability to control style, with the potential benefit of helping with mixing  and meaning preservation.  We adapt a part-of-speech tagger proposed by ~\citet{sajjad-NAACL} and tag teach token in our initial sequence. While sampling an edit point, we do not allow the model to sample a token which is tagged as any form of noun. The improvement from disallowing any edits to nouns can be observed in our style transfer results~(\autoref{tab:style-transfer}).

\section{Style Control Experiments}\label{sec:style-control-experiments}

\subsection{Metrics} \label{sec:metrics}

Discriminating between fine-trained styles (i.e. on a per author basis) is a challenging task for human evaluators. Therefore, our evaluation of control success relies on automatic metrics. To avoid concerns about gaming certain metrics, we include multiple automatic metrics for each text attribute that is measured. We measure the overall quality of the generated text through fluency in addition to particular features (e.g. semantic meaning, style consistency) in generated text \cite{Celikyilmaz2020EvaluationOT}. We also consider further downstream tasks to evaluate the quality of style altered text, like author detection (\autoref{sec:anonymization}) and LM generated text detection (\autoref{sec:detection}).
\vspace{-5pt}
\paragraph{Style similarity} To measure how well generated text matches a target style, we adapt previous work (discussed in \autoref{sec:prelim}) as automatic evaluation tools. We consider ``Universal Author Representations'' (UAR) \cite{luar} and ``Content Independent Style Representations'' (CISR)~\cite{wegmann-etal-2022-author}\footnote{Checkpoints: \url{https://huggingface.co/rrivera1849/LUAR-CRUD} and \url{https://huggingface.co/AnnaWegmann/Style-Embedding}.}.
These pre-trained embeddings measure style overlap between generated and reference text samples. We report cosine similarity between reference and generated text embeddings.

\paragraph{Fluency} Beyond satisfying style-specific constraints, generated text should remain fluent. We emphasize however that fluency is, to some extent, at odds with the goal of introducing author-specific style.\footnote{The human reference data in \autoref{tab:text_control_main} has an average perplexity of 205.04 measured by Mistral-7B.} That is, if the target style samples have a high perplexity under an LM, it is reasonable to expect a well formed style controlled generation to also have a high perplexity. We directly measure and report the percent difference between generated and reference fluencies throughout our results, where a smaller difference is better. We use Mistral-7B to measure fluencies of all generated text \cite{jiang2023mistral}. 

\subsection{Experimental Setup}\label{sec:style-details}

\begin{figure}
\centering
  \includegraphics[width=3.0in]{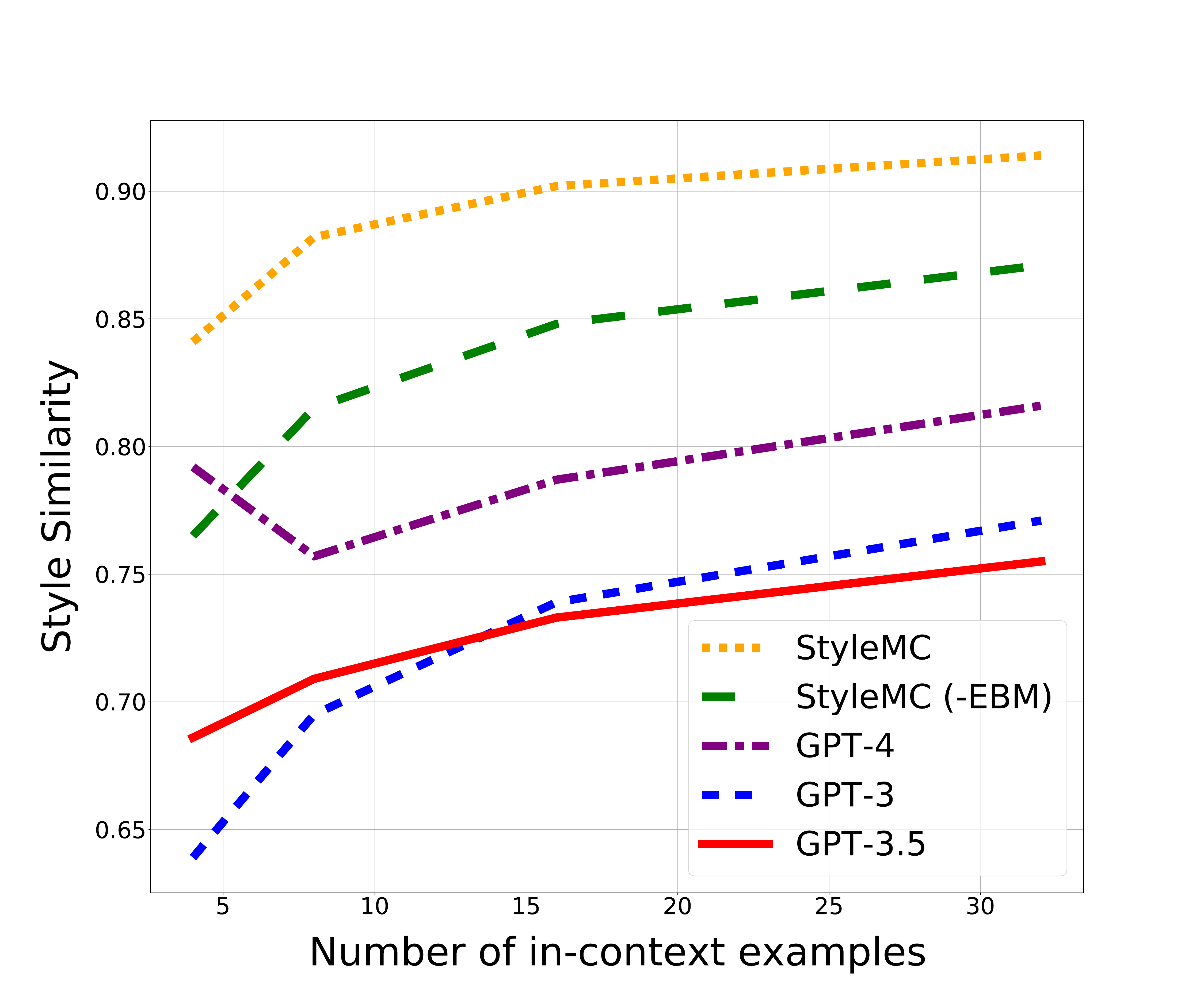}

\caption{Style performance for an increasing number of examples of a target style. We find that more examples result in better representations, which in turn improve decoding quality. Our proposed  approach, including an ablation without sequence-level decoding (-~EBM), significantly outperforms much larger models using prompting strategies.}

\label{fig:main_incontext}
\end{figure}
\paragraph{Evaluation datasets}\label{sec:style-datasets}
We evaluate the effectiveness of our decoding strategy on 800 authors contributing to four unique Reddit subreddits. We consider Reddit data collected through the Pushshift API \cite{baumgartner2020pushshift} and compile a test split for each of our 4 subreddits: \texttt{/r/wsb}, \texttt{/r/AskHistorians}, \texttt{/r/news}, and \texttt{/r/australia}; performance on all four results are reported together in \autoref{tab:text_control_main}. These subreddits are selected for their unique and distinctive styles. We use an additional split from \texttt{/r/wsb} to validate our methods (i.e. to select optimal decoding hyperparameters). For each author, we compile $N$ text samples as the source of style evidence. In all experiments except for Figure \ref{fig:main_incontext}, we set $N=16$ to balance the quality of resulting style representations and cost of compiling data following previous work \cite{andrews-bishop-2019-learning}. Figure \ref{fig:main_incontext} illustrates part of this trade-off, with improved style controlled generation for larger $N$ values.

\vspace{-5pt}\paragraph{Language models} Our proposed methods operate on a frozen underlying LM; we do not perform any fine-tuning. Across all experiments we use variants of OPT and MPT-7B to generate text \cite{OPTarxiv, MosaicML2023Introducing}.\footnote{A single NVIDIA V100 GPU was sufficient to store both the control model and language model, except for 7B parameter experiments which required a second V100 GPU.} For EBMs, we use T5-3B as our proposal distribution.

\vspace{-5pt}\paragraph{Future Regressors}
We train the forward looking regressor using a single V100 GPU, a batch size of 64, and a learning rate of $1e^{-6}$ for 100k steps.
In our experiments we apply this re-scoring procedure to OPT-1.3B
to capture author specific fluencies, and we evenly weight likelihoods from OPT and the forward regressor to compute re-scored sequence level likelihoods. 

\begin{table*}[ht]
    \centering
    \begin{tabular}{lrccc}
    \toprule

       & & \multicolumn{2}{c}{$\uparrow$ Success} \\
       \cmidrule(lr){3-4}
       Model & Size & UAR & CISR & $\downarrow$$\Delta$ Fluency   \\
       
    \midrule
       Human Reference & &  0.852 &  0.864 & \\
    \midrule
        GPT-3  & 175B& 0.755 &  0.649 & 110.2\% \\
      GPT-3.5 & &   0.742 &  0.560 & 77.23\%  \\
      GPT-4 & & 0.767 & 0.637 & 46.93\% \\

    \midrule
      
       \multirow{4}{*}{Future Regressor (OPT)} & 350M + 82M   & 0.791 &  0.727 &  706.14\%   \\
        & 1.3B + 82M &  0.789 &  0.707 &  625.97\%  \\
        & 2.7B + 82M & 0.759 &  0.730 & 128.92\%   \\
        & 6.7B + 82M & 0.767 &  0.721 &  78.38\% \\
       Future Regressor (MPT) & 7B + 82M& 0.788 &  0.715 & 95.33\% \\
    
        MuCoLa \cite{kumar2022gradientbased} & 125M + 82M &  0.709 & 0.596 & 677.24\% \\
        \midrule
        \name \hspace{2pt}(Revise Future Regressor)& 3B + 1.3B + 82M&  \textbf{0.849} &  \textbf{0.762} & 322.33\%  \\
        \name \hspace{1pt} (Revise GPT-4)& &  0.838 &  0.736 & \textbf{39.87\%}  \\
        
    \bottomrule
    \end{tabular}
    \caption{Test results on all four subreddit test splits. The proposed future regressor approach outperforms both prompting approaches on the target control metric (UAR) and the secondary style metric (CISR). Both our proposed model and MuCoLa revise the output of the OPT-350M future regressor model. We additionally use our method to revise GPT-4 outputs, finding significant improvements. In both cases, revising outputs with our method achieves the highest performance, approaching the human reference. We omit model sizes for GPT-3.5 and GPT-4, as they are unknown. A paired sign test of the differences between our proposed method and GPT-4 is significant at at least the $p < 10^{-7}$ level for all metrics.}

    \label{tab:text_control_main}

\end{table*}

\vspace{-5pt}
\paragraph{Baselines}\label{sec:prompted_lms}
We use few-shot prompting with GPT-3~\cite{brown2020language}, GPT-3.5, and GPT-4. 
Since stylistic generation does not impose a constraint on semantic similarity, we found that a simple prompt was sufficient, compared to the schemes proposed by~\citet{reif-et-al} and \citet{patel2022lowresource} for style transfer.
We provide a template for our prompt, where each writing sample in the prompt is truncated to 32 tokens.
\begin{small}
\begin{verbatim}
    Here are some passages of text:
    <author writing sample 1>
    <author writing sample 2>
    ...
    <author writing sample 16>

    Write another passage in the
    same style:
\end{verbatim}
\end{small}

For the GPT-3 baselines, we use the largest model, Davinci, with 175 billion parameters \cite{brown2020language}; for GPT-3.5 baselines, we use the \texttt{gpt-3.5-turbo-0613} snapshot; for GPT-4 we use \texttt{gpt-4-0613}. For all models, we use a temperature of 1.0 and frequency penalty of 2, terminating generations after 32 tokens.

We also compare our proposed text revision approach to MuCoLa \cite{kumar2022gradientbased}, a sampling procedure that uses gradients to optimize over differentiable constraints. We report the results in \autoref{tab:text_control_main}. A limitation of this approach is that all models must share the same embedding table; we use a LUAR style model trained to use the OPT-125M embedding table, which obtains within 4.7\% of the performance of the style model. We construct an energy function with the goal of maximizing the cosine similarity between the target style embedding and the style embedding of a single output from our future regressor. We tune hyperparameters on the validation split of our dataset and use the best hyperparameters reported for the sentiment task in the original paper with several exceptions: we use the weighted sum selection criteria with a weight of 0.001 on the language model and 0.999 on the style model, a threshold of -5, and maximum length of 32 tokens.

\subsection{Style-Controlled Generation} \label{sec:style-ctrl-gen}
Table~\ref{tab:text_control_main} compares the proposed approach to ablations consisting of just the adapted LM using future regressors, and prompting-based methods using state of the art LMs. The results in the final two rows use our author-adapted LMs to sample initializations for revision.\footnote{We did experiment with revising GPT outputs as well, but found that revising the future regressor output yielded better results. This initialization is supported by our use of future regressors to balance fluency in our energy function.} The first row shows metrics for ``gold" style matches, i.e. additional held out text samples written by the \emph{same} human author are used for comparison. Our proposed decoding strategy performs competitively despite the fact that the baseline LMs are much larger and have undergone steps like instruction tuning in the case of GPT-3.5~\cite{instructgpt}. Under the UAR style metric, our proposed future regressor method outperforms baseline LMs using in-context learning. When the outputs of the future regressor are revised using the EBM text revision method described in \autoref{sec:ebm}\footnote{For these experiments, we iteratively sample for 5 epochs, where an epoch iterates for the number of tokens in the longest sentence in the batch.}, it outperforms the prompting method on both success metrics, also shown in Table \ref{tab:text_control_main}. We additionally demonstrate that~\name~can improve the outputs of prompted LMs; we significantly improve control results by revising the outputs of GPT-4. Across all of our methods, we find that the amount of stylistic evidence made available directly impacts decoding performance (Figure \ref{fig:main_incontext}).

\subsection{Style Vector Interpolation}\label{sec:interpolation}
\begin{figure*}[t]
\centering
\begin{subfigure}
  \centering
  \includegraphics[width=2.5in]{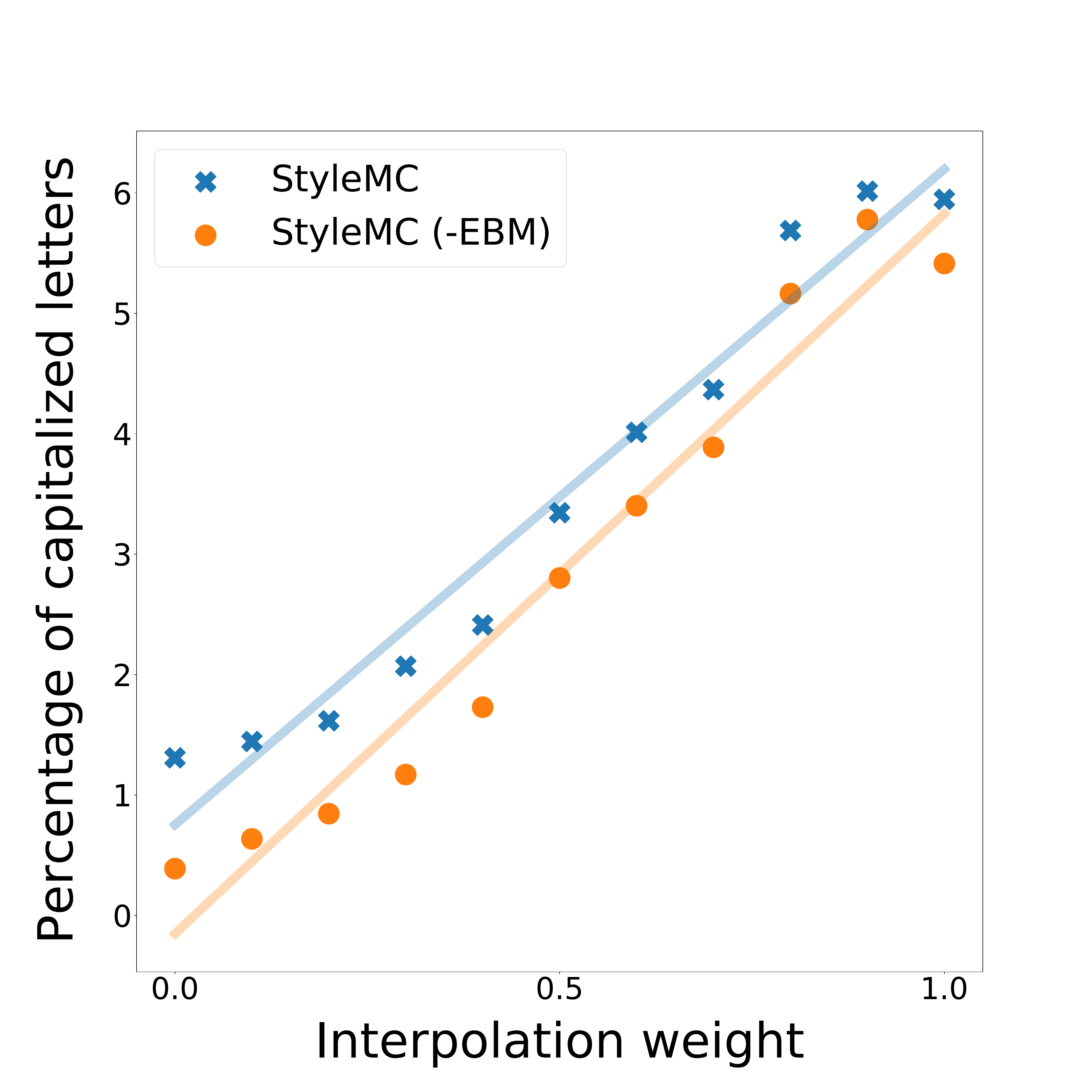}
  \label{fig:plot_caps}
\end{subfigure}%
\begin{subfigure}
  \centering
  \includegraphics[width=2.5in]{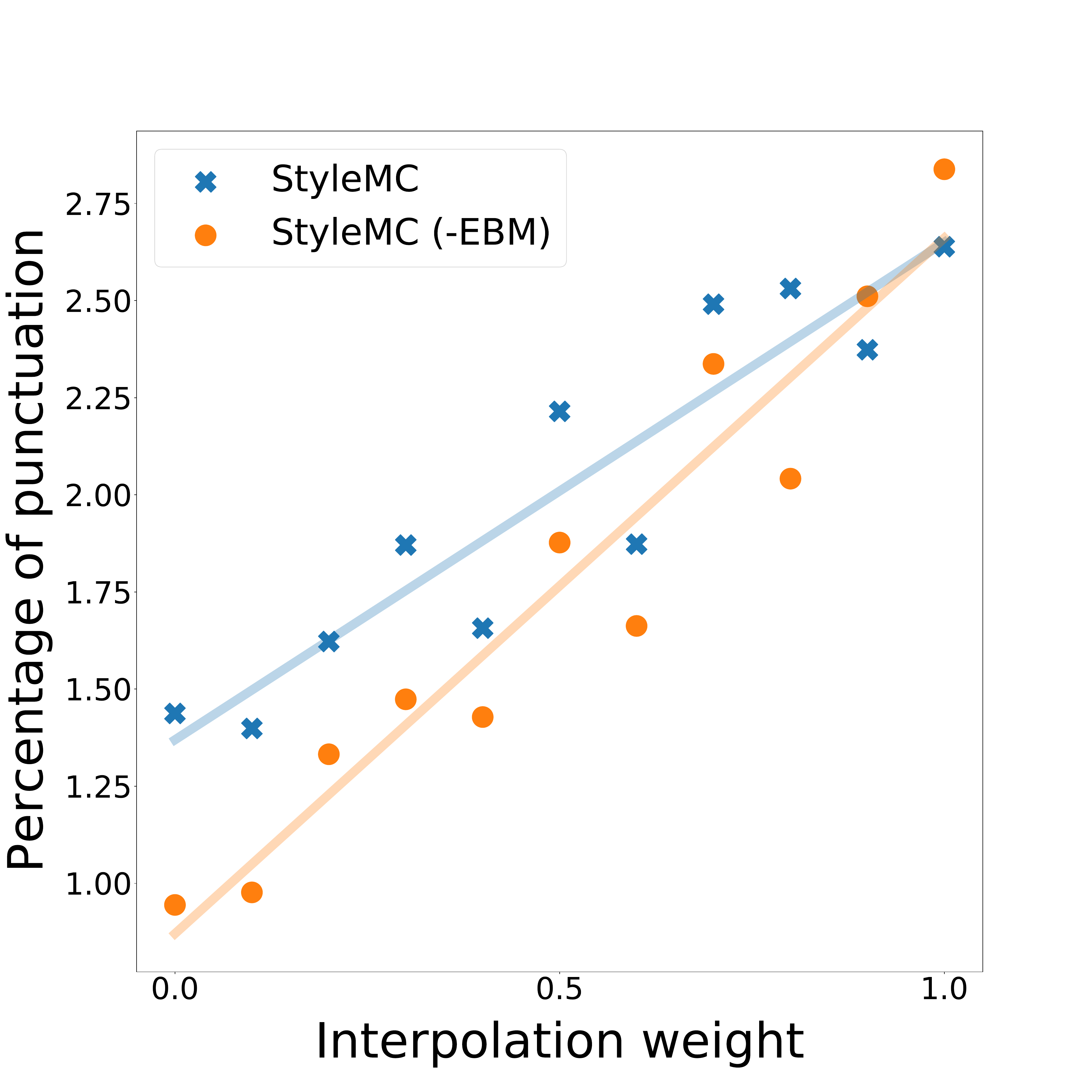}
  \label{fig:plot_punct}
\end{subfigure}
\caption{Percent of capitalized and punctuation characters in generated outputs. The decoding procedure is run on \emph{interpolated} style vectors, where a weight of 0.0 is indicates a style vector capturing a \texttt{nocaps} or \texttt{nopunct} behavior, and a weight of 1.0 corresponds to a normal \texttt{/r/wsb} user.}

\label{fig:interpolation}
\end{figure*}

We construct two artificial datasets with known stylistic attributes: \texttt{nocaps}, composed of data from 25 users of \texttt{r/wsb} converted to only lowercase characters, and \texttt{nopunct} composed of data from 25 users of \texttt{r/wsb} with all punctuation removed. We select these two attributes because they are easy to qualitatively identify and illustrate specific levels of control. We generate the UAR embedding for each author in \texttt{nocaps} or \texttt{nopunct} and interpolate it using spherical geometric interpolation\footnote{Specifically, we use the \texttt{scipy} implementation of spherical geometric interpolation.} with the UAR embedding for the same 25 authors in \texttt{r/wsb} with varying weights. We generate outputs using the future regressor, and further modify these outputs using the EBM. We find that stronger bias towards the \texttt{nocaps} UAR embedding tends to measurably decrease the amount of capital characters in the text and that stronger bias towards \texttt{nopunct} measurably decreases the amount of punctuation in generated text (\autoref{fig:interpolation}), demonstrating that both models can replicate meaningful features of style encoded by the UAR control vectors.

\section{Style Transfer Experiments}\label{sec:style-transfer-section}
In our second experiment, building on our style control experiments we explore whether we can produce text in an arbitrary writing style while preserving the meaning of the original text.  
To control both style and meaning, we add a semantic similarity expert in the form of an SBERT encoder as described in \autoref{sec:ebm}.

\subsection{Metrics}

The same metrics used to evaluate style control in \autoref{sec:metrics} can be used for style transfer. However, since style transfer necessitates preservation of meaning, we introduce another metric to measure this.

\vspace{-5pt}
\paragraph{Semantic similarity}

Since our experiments involve short documents, we consider semantic search models which provide a document-wide notion of semantic similarity. Specifically we employ (1) \texttt{all-mpnet-base-v2}, a high-performance SBERT model and (2) GTR \cite{gtr}, a large dual encoder trained for semantic search. Note that the SBERT model used for evaluation is distinct from the SBERT model used to guide generation (\texttt{all-MiniLM-L6-v2}).

\vspace{-5pt}
\paragraph{Edit distance}
To convey how much a sequence is changed to achieve a target style, we measure a Levenshtein distance between the initial text and final outputs from each method. Intuitively, it is desirable to make parsimonious edits to achieve a desired style, such as in the anonymization experiments in~\autoref{sec:anonymization}.

\subsection{Experimental Setup}

For style transfer we create a dataset pairing Reddit comments with arbitrary target styles, where each target style consists of 16 comments from the same Reddit user and Subreddit. We select three author styles at random from each of the 4 subreddits specified in \S\ref{sec:style-datasets} and \texttt{/r/casualUK} for a total of 15 target styles. Since our author styles are derived from the author's comment history, we can pair these comments with the other author styles in a round-robin manner. We exclude pairings between comments and styles that co-occur in the same subreddit, yielding a total of $2880$ pairs.
For EBMs, as proposed in \autoref{sec:ebm} we include a ``meaning preservation'' expert. For this expert, we use the \texttt{all-MiniLM-L12-v2} Sentence Transformers model.

\vspace{-5pt}
\paragraph{Prompting Baseline}

For in-context style transfer, we use a 2-shot variation of the approach described by \citeauthor{patel2022lowresource}, where text is first paraphrased into a neutral style before being rewritten to match the target style. We keep the same hyperparameters as in \autoref{sec:prompted_lms}, except that we extend the maximum length of generations to 64.

\subsection{Style Transfer Results}

\begin{table*}[ht]
    \centering
    \begin{tabular}{l|cccccc}
    \toprule
       & \multicolumn{2}{c}{\small $\uparrow$ Success} & \multicolumn{2}{c}{\small $\uparrow$ Similarity} & \\
       \cmidrule(lr){2-3} \cmidrule(lr){4-5}
       \small Model & \small UAR & \small CISR & \small SBERT & \small GTR & \small $\downarrow$ $\Delta$Fluency & \small $\downarrow$ Levenshtein \\
    \midrule
    \small Initial State & 0.660 & 0.435 & 0.042 & 0.438 & 781.78\%& \\
    \midrule
    \small GPT-3 & 0.733 & 0.749 & 0.650 & 0.784 & 711.69\% & 71.68\\

    \small GPT-3.5 & 0.701 & 0.590 & 0.760 & 0.853 &  \textbf{147.76\%}  & 64.52\\

    \small GPT-4 & 0.738 & 0.723 & 0.741 & 0.841 &  742.31\% & 62.60 \\
    \midrule
    \small \name & 0.768 & 0.618 & \textbf{0.968} & \textbf{0.978} & 701.21\% & \textbf{14.71}\\
    \small \name \hspace{1pt}(\small GPT-4 Initialization) & \textbf{0.791} & \textbf{0.802} & 0.728 & 0.832 & 735.98\% & 17.83\\
    \midrule
    
    \hspace{0.25cm}(\small \emph{-Noun Constraints}) & 0.761 & 0.583 & 0.963 & 0.971 & 807.43\% & 23.21\\
    \hspace{0.25cm}(\small \emph{-Style Expert}) & 0.662 & 0.432 & 0.991 & 0.993 & 794.84\% & 1.91\\
    \hspace{0.25cm}(\small \emph{-Semantic Expert}) & 0.831 & 0.680 & 0.501 & 0.702 & 1013.11\% & 47.67\\
    \hspace{0.25cm}(\small \emph{-Fluency Expert}) & 0.755 & 0.599 & 0.968 & 0.976 & 803.98\% & 14.79\\

    \bottomrule
    \end{tabular}
    \caption{Test results for style transfer. The EBMs revise human written text to the style of a different author from a different subreddit. We provide metrics for the initial text, before style transfer, as a reference point. The last column represents a Levenshtein edit distance from the model outputs to the initial text. A paired sign test of the differences between our proposed method and GPT-4 is significant (at least) at the $p < 10^{-7}$ level for all metrics. We bold the best score for each metric, ignoring ablations.} 
    \label{tab:style-transfer}
\end{table*}

In Table \ref{tab:style-transfer}, for style-transferred text produced by each method, we report the extent to which the target style is achieved (UAR, CISR), fluency as measured by GPT-2, and the extent of semantic preservation (SBERT, GTR). As a trivial baseline, we report the same metrics between pairs of unrelated text samples, which was achieved by comparing the style reference to the initial content (i.e. initial state).
Our approach performs comparably to prompted large language models while requiring only a fraction of the number of parameters. 

While searching for hyperparameters, we observed a trade-off between stylistic accuracy and content preservation. This observation is consistent with the notion that style and content cannot be completely disentangled. For instance, optimizing for content preservation may introduce stylistic features from the target content into generated text, and vice versa. We handle this trade-off by providing hyperparameters for tuning the relative importance between semantic preservation and style accuracy, providing a ``knob'' to tune for different applications. We show qualitative results in \autoref{tab:success_qualititative} and \autoref{sec:appendix_qual}, demonstrating that prompted LMs often fail to capture finer-grained changes necessary for style transfer. Our proposed models are able to selectively edit sequences to satisfy these constraints.

\begin{table*}[h]
    \centering{
    \begin{tabular}{p{3.2 cm}|p{12cm}}
    
    \hline \hline
        Target Style & "Reuse requires more effort/\colorbox{Goldenrod}{labour}. People are lazy, especially in australia."\\
        & "What you learn at \colorbox{Goldenrod}{uni} isnt \colorbox{Goldenrod}{rubbish} but really a framework for your skills. Essentially you're learning to learn."
        \\
        & "I'd like Tanya too but in general Australian just isn't ready a female leader again."
        \\
        \hline \hline
         Initial Content & "lol global warming is a fact? Please, we all know it's not after they changed the goal post to `climate change' instead"\\
         & "this brings me some hope for my -70\% gains."\\
         & "so are you saying we aren't using atrazine as herbicide for our crops?" \\
         \hline \hline
         GPT-4 & "The phrase `global warming' was rebranded as `climate change', which caused some to doubt its authenticity." \\
         & "This brings a glimmer of hope to my 70\% losses." \\ 
         & "Are you saying we're not using atrazine as a pesticide on our crops?" \\
         \hline \hline
         Proposed\hspace{2pt}(RoBERTa Proposals)& "\textcolor{red}{Our climate} warming\textcolor{red}{'s not} fact. \textcolor{red}{Worse, others have \colorbox{Goldenrod}{realised} it is not because we switched} the goal post to `climate change' instead" \\
         &"\textcolor{red}{That gives} me some hope for my -70\% gains \textcolor{red}{today.}" \\
         & "\textcolor{red}{Why were we} saying we aren't \textcolor{red}{doing} atrazine as herbicide for \textcolor{red}{the crops.}" \\

         \hline \hline
         Proposed\hspace{2pt}(T5-3B Proposals)& "\textcolor{red}{To that end. If} global warming is a fact? \colorbox{Goldenrod}{\textcolor{red}{Ahhh, bloody heck,}} we all know it's not after they \textcolor{red}{had switched} the goal post to just `climate change' instead" \\
         &"\textcolor{red}{The above has given} me some hope for my -70\% gains." \\
         & "\textcolor{red}{S}o are you saying that we aren't using atrazine as herbicide for our crops?" \\
    \bottomrule
    \end{tabular}}
    \caption{Successful style transfer from a \texttt{r/wsb} author to a \texttt{r/australia} author. Obvious stylistic behavior is highlighted in orange, and edits by our proposed models are in red font. GPT-4 is able to edit punctuation and capitalization, but fails to capture finer-grained features. T5-3B is able to replace multiple tokens at a time when necessary, as indicated in this table leading to better style transferred results.}
    \label{tab:success_qualititative}

\end{table*}

\section{Additional Experiments}\label{sec:added-experiments}
\subsection{Anonymization} \label{sec:anonymization}Enabling author privacy is a promising application of \name. Previous work has explored the preservation of privacy by altering identifying linguistic features associated with text \cite{Li2018TowardsRA}. We measure success by the system's ability to circumvent an author attribution system. In this setting, author attribution involves attempting to match text samples $Q$ (queries) and $T$ (targets) that were written by the same author. We consider a subset of the authors in the Reddit dataset to evaluate attribution capabilities. Our sample consists of 180 authors, and results in 32,400 binary comparisons. Given a user's history made up of $N$ posts, we take the first $N/2$ posts and establish a query ($Q)$, and the second $N/2$ posts to establish a target ($T$). In our experiments we use $N = 16$. Using our proposed style transfer approach, we alter the style of each target $T$ to produced a perturbed target $T'$. Success is measured by the decrease in performance when matching $Q \rightarrow T'$ compared to $Q \rightarrow T$.
To evaluate our approach, we consider all possible pairs of queries and targets and seek to detect matching queries and targets before and after style transfer is applied. We extract representations using UAR  for each query and target sample, and compute pairwise distances to use as scores \cite{luar}. A smaller score in this case indicates a higher likelihood that the two representations are from the same author. Solving the detection problem involves setting an operating point with a given rate of false positives and false negatives, the point at which the two rates are equal is known as the equal error rate. A lower value indicates a better detection result. The results in Table \ref{tab:style-transfer-anon} show that our procedure successfully reduces the detection rate through style transfer.
\begin{table}[ht]
    \centering
    \begin{tabular}{l|cc}
    \toprule
       Model & EER$_{\text{before}}$ $\rightarrow$ EER$_{\text{after}}$  & $\downarrow$Lev \\
    \midrule
    GPT-3 & 0.116 $\rightarrow$ 0.311 & 71.68 \\
    GPT-3.5 & 0.116 $\rightarrow$ 0.278 & 64.52 \\
    GPT-4 & 0.116 $\rightarrow$ 0.324 & 62.60\\
    \midrule
    Proposed  & 0.116 $\rightarrow$ \textbf{0.363} & \textbf{14.71}\\
    \bottomrule
    \end{tabular}
    \caption{Extent of anonymization after style transfer. An increasing EER via style transfer indicates improved anonymization. The last column is a Levenshtein distance between the model output and original text. \name \hspace{3pt}is able to more effectively anonymize writing with fewer edits.}
    \label{tab:style-transfer-anon}
\end{table}

\subsection{Detection of generated text}\label{sec:detection}

Considering the potential for misuse of generative text, especially in the context of style control, we conduct a small study on the detectability of our proposed future regressor decoding strategy. We find that similar to popular LMs like GPT-3, detecting text from our method in a \emph{zero-shot} setting is quite difficult, with a classifier incorrectly marking fake text as human-written with high confidences. However given a relatively small set of examples (in our experiments we consider 500 samples of generated text from each LM), detection of LM generated text becomes more tractable with basic classification approaches.

\begin{table}[h]
    \centering
\begin{tabular}{l|cc}
    \toprule
        Strategy & GPT-3 & \name   \\
    \midrule
       Zero-shot & 0.676 &  0.547 \\
       In-domain Training & 0.972 &  0.831 \\
    \bottomrule
    \end{tabular}
    \caption{Detection accuracy for text sampled from GPT-3 and our proposed decoding strategy. Each split consists of 250 real and 250 fake text samples.}
    \label{tab:faketext}
\end{table}

To construct a dataset for this task, we follow a strategy used by OpenAI's fake text detector \cite{AITextClassifier}. Similar to our main experiments, we use the Pushshift API to collect real text samples from 10,000 Reddit users, ensuring that each sample has at least 16 posts \cite{baumgartner2020pushshift}. We concatenate this data to create a prompt, and allow OPT-6.7B to generate follow on fake text for the prompt. The resulting dataset consists of 10,000 human written text samples and 10,000 machine generated outputs associated with those prompts. Additionally, we construct two more datasets which include 500 GPT-3 samples and 500 samples from our proposed EBM strategy to demonstrate improved detectability when in-domain data is considered. We fine-tune a RoBERTa base model \cite{liu2019roberta} on these datasets for 10 epochs on a single V100 GPU using a learning rate of 2e-5 and AdamW optimizer \cite{loshchilov2019decoupled}.

 \autoref{tab:faketext} shows test accuracy on a subset of the test data (strictly \texttt{/r/wsb} users) used in our main experiments (Table \ref{tab:text_control_main}). In both cases, performance is quite poor in the zero-shot setting. When in-domain training data is considered, text sampled from GPT-3 is detected at a high rate. We note that fake text from our proposed strategy is detected at a significantly higher rate compared to the zero-shot setting, but not nearly as high as GPT-3. This is likely due to the perturbations applied to the LM distribution by the proposed method. While lower detection accuracies are a good result for style-control, it does raise misuse concerns. Our result also shows that these concerns can be balanced if more in-domain text is available, increasing the rate of detection of style-revised text.

\section{Related Work}\label{sec:related_work}

Effective text style transfer is important for many downstream applications such as writing assistants, personalized NLP systems, text simplification, detoxifying and debiasing text \cite{jin-etal-2022-deep}. Interest in the task has led to many datasets spanning various types of styles and domains \cite{briakou-etal-2021-ola, madaan-etal-2020-politeness, rao-tetreault-2018-dear} and approaches \cite{prabhumoye-etal-2018-style, krishna2020reformulating, riley2021textsettr, Hallinan2023STEERUS}. However, these approaches largely focus on \emph{coarse} level styles (e.g. formality, politeness, simplicity) rather than \emph{fine-grained} styles which may contain any combination of coarse styles. For finer-grained style transfer, \citet{riley2021textsettr} propose a few-shot strategy using learned style vectors to autoregressively decode text. Our work differs by using a pre-existing encoder for style vectors \cite{luar} and incorporating \textit{bidirectional} context during inference. In high resource settings, parallel corpora can be leveraged to directly learn relationships between styles \cite{jhamtani-etal-2017-shakespearizing}, however having such datasets for arbitrary authors is not realistic. Additionally, recent interest in prompting large language models has facilitated style transfer from arbitrary authors using in-context learning \cite{reif-et-al, patel2022lowresource}. We use similar prompting strategies as comparable baselines for our approach. 

Research towards controllable text generation has focused on fine-tuning approaches, discriminator guided decoding, and more recently on large-language model prompt engineering. Fine-tuning approaches condition a language model on a given control attribute. For a control attribute $c$, the language model is trained to predict the probability of the next word $p(x \mid c)$. This probability can be directly modelled as in the case of CTRL which uses an initial control prefix to guide decoding \citep{keskar2019ctrl}. However, CTRL requires re-training a LM any time a new control code is proposed. One way to avoid training from scratch is to approximate the probability $p(x \mid c)$ as $p(c \mid x)p(x)$. Here $p(x)$ can be modeled by a pre-trained language model and $p(c \mid x)$ can be modeled by a simple discriminator. Rather than training an entire language model, only the discriminator would need to be trained \citep{dathathri2019plug, krause2020gedi, yang2021fudge}. However, as noted in~\autoref{sec:intro}, control attributes perform poorly on finer-grained tasks, motivating the use of control \textit{vectors} instead.

\section{Conclusion}\label{sec:conclusion}
With \name~we have demonstrated the ability to guide the style of generated text using author representations, which capture fine-grained aspects of writing style on the basis of a small writing sample. We develop a novel sequence-level model for this purpose, consisting of an author-adapted LM and non-autoregressive inference procedure. The proposed approach outperforms large intruction-tuned LMs at guiding generated text towards the desired attributes.

\vspace{-5pt}\paragraph{Limitations} The main limitation of our study is the reliance on automatic evaluation metrics. To avoid relying on any single automatic metric, we include a diverse set of evaluation strategies, particularly the interpolation experiments in~\autoref{sec:interpolation} that focus on \emph{interpretable} stylistic attributes. The success of the interpolation experiments provide support  both the effectiveness of the style representations as well as our ability to generate text in the target style. In the case of coarse style transfer problems like formality and sentiment, non-expert human annotators can perform the task and therefore be used to complement automatic metrics. However, authorship attribution requires trained forensic linguists, an avenue which we decline to pursue in this work, both for cost reasons and to avoid setting a precedent that may detract from future work in this area. Similar to previous efforts in controllable generation, the proposed approach uses a discriminative model to guide generation, and success at control is reliant on the quality and availability of appropriate training data to estimate that model. In our experiments, we rely on representations of author style that are trained on large amounts of anonymous social media content and are highly discriminative of authorship~\cite{luar}. However, social media data may contain various biases, such as a prevalence of English over other languages, as well as biases owing to the sample sizes of various demographic groups relative to the population.

\vspace{-5pt}\paragraph{Broader Impact} This paper pushes the state of the art in style-controlled text generation, which enables a number of downstream applications, such as writing assistants, anonymization (e.g., for political dissidents), and personalized NLP more broadly, such as for under-represented groups. Another interesting potential application area is machine-translation, tailored to an author's particular style.  We are also excited about potential applications of style-controlled generation to data augmentation and synthetic data creation with LLMs, which may otherwise suffer from lack of diversity relative to real data composed by a variety of authors with distinct styles.  At the same time, as with most technologies there is potential for abuse. In \autoref{sec:added-experiments}, we address one way methods discussed here may be abused: defeating machine-text detectors. We explore a mitigation scheme involving retraining the detector on style-controlled outputs (\autoref{tab:faketext}), showing that this results in drastic improvements in detection accuracy. Few-shot detection approaches could also be effective in mitigating abuses of the proposed method~\cite{soto2024fewshot}.

\section*{Acknowledgements} This research is supported in part by the Office of the Director of National Intelligence (ODNI), Intelligence Advanced Research Projects Activity (IARPA), via the HIATUS Program contract \#D2022-2205150003. The views and conclusions contained herein are those of the authors and should not be interpreted as necessarily representing the official policies, either expressed or implied, of ODNI, IARPA, or the U.S. Government. The U.S. Government is authorized to reproduce and distribute reprints for governmental purposes notwithstanding any copyright annotation therein.

\bibliographystyle{acl_natbib}
\bibliography{tacl2021,anthology,custom}

\appendix
\section{Additional Qualitative Results} \label{sec:appendix_qual}
\autoref{tab:success_qualititative2} demonstrates additional style transfer model outputs, here the transfer occurs in the opposite direction of \autoref{tab:success_qualititative}. In one case, both of our proposed models fail to make any edits to the initial text during the sampling procedure. We again observe GPT-4 producing fluent and readible text, but with no noticeable style specific edits.
\begin{table*}[h]
    \centering{
    \begin{tabular}{p{3.2 cm}|p{12cm}}
    
    \hline \hline
        Target Style & "\colorbox{Goldenrod}{omg} thats amazing\colorbox{Goldenrod}{!! w}here'd it go? do it again"\\
        & "\colorbox{Goldenrod}{t}his brings me some hope for my -70\% gains"
        \\
        & "\colorbox{Goldenrod}{lol} global warming is a fact? Please, we all know it's not after they changed the goal post to `climate change' instead"
        \\
        \hline \hline
         Initial Content & "Reuse requires more effort/labour. People are lazy, especially in australia."\\
        & "What you learn at uni isnt rubbish but really a framework for your skills. Essentially you're learning to learn."
        \\
        & "I'd like Tanya too but in general Australian just isn't ready a female leader again." \\
         \hline \hline
         GPT-4 & "recycling takes work. aussies especially are too lazy to do it cause they want things easy and convenient." \\
         & "the knowledge you learn in college isn't worthless, it gives you a foundation of skills. essentially, you're getting schooled on how to school yourself." \\ 
         & "i like Tanya but aus ain't ready for another sheila in charge." \\
         \hline \hline
         Proposed\hspace{2pt}(RoBERTa Proposals)& "Reuse requires more effort/labour. People are lazy, especially in australia." \\
         &"\textcolor{red}{w}hat you learn at uni isnt rubbish \textcolor{red}{sonz} a framework \textcolor{red}{to} your skills. Essentially you learned how to learn more" \\
         & "\textcolor{red}{i}'d like Tanya too\textcolor{red}{; good fact} Australian just isn't ready a female leader again." \\
         \hline \hline
         Proposed\hspace{2pt}(T5-3B Proposals)& "Reuse requires more effort/labour. People are lazy in australia." \\
         &"What you learn at uni isnt rubbish but really a framework for your skills. Essentially you're learning to learn." \\
         & "\colorbox{Goldenrod}{\textcolor{red}{tf}} I like Tanya \colorbox{Goldenrod}{\textcolor{red}{so much now...}} but \textcolor{red}{I really can't believe she's a feminist. Australian politics is never going to have a single male} leader again." \\
         
    \bottomrule
    \end{tabular}}
    \caption{Style transfer from a \texttt{r/australia} author to \texttt{r/wsb} author. Obvious stylistic behavior is highlighted in orange, and edits by our proposed models are in red font. GPT-4 paraphrases the outputs but does not effectively transfer style.}
    \label{tab:success_qualititative2}

\end{table*}

\section{Proposal Model Variations}\label{sec:proposal_appendix}
We experiment with several different components of our proposed model, which masks a portion of the current sequence, and samples an alternative infill. We find that the ability to produce variable length proposals to be crucial to producing high quality outputs. \autoref{tab:proposal_models} demonstrates the significantly better job done by T5 at matching reference fluencies, this can be qualitatively observed in \autoref{tab:success_qualititative} as well. We also vary the the masking procedure, finding that sampling two tokens at a time produces the best results (\autoref{tab:masking_window}).

\begin{table}
    \centering
    \begin{tabular}{l|ccc}
    \toprule
       \small Model & \small $\uparrow$ UAR & \small $\uparrow$ SBERT &  \small $\downarrow$ $\Delta$Fluency  \\
    \midrule
    \small Initial State & 0.674  & 0.032 & 792.73\%  \\
    \midrule
    RoBERTa-Base & 0.766  & 0.949 & 1801.01\% \\
    T5-base & 0.746  & 0.969 & 791.85 \\
    T5-large & 0.752  & 0.965 & 807.21\% \\
    T5-3B & 0.768  & 0.968 & 701.21\% \\
    \bottomrule
    \end{tabular}
    \caption{Style transfer results for varying proposal models. RoBERTa proposes single token edits, while T5 models may propose variable edits.}
    \label{tab:proposal_models}

\end{table}

\begin{table}
    \centering
    \begin{tabular}{l|ccc}
    \toprule
       \small Model & \small $\uparrow$ UAR & \small $\uparrow$ SBERT &  \small $\downarrow$ $\Delta$Fluency  \\
    \midrule
    \small Initial State & 0.674  & 0.032 & 792.73\%  \\
    \midrule
    Mask$_{1}$ & 0.750  & 0.966 & 803.82\%\\
    Mask$_{2}$ & 0.761  & 0.968 & 701.21\% \\
    Mask$_{3}$ & 0.749  & 0.966 & 755.74\% \\
    
    \bottomrule
    \end{tabular}
    \caption{Style transfer results for varying masked window sizes in our T5 proposal model.}
    \label{tab:masking_window}
\end{table}

\end{document}